# Systematic Evaluation of Preprocessing Techniques for Accurate Image Registration in Digital Pathology

**Fatemehzahra Darzi** [1,2], **Rodrigo Escobar Díaz Guerrero** [1,2], and **Thomas Bocklitz** [1,2,*]

[1] Institute of Physical Chemistry (IPC) and Abbe Center of Photonics (ACP), Friedrich Schiller University Jena, Member of the Leibniz Centre for Photonics in Infection Research (LPI), Helmholtzweg 4, 07743 Jena, Germany

[2] Department of Photonic Data Science, Leibniz Institute of Photonic Technology, Member of Leibniz Health Technologies, Member of the Leibniz Centre for Photonics in Infection Research (LPI), Albert-Einstein-Strasse 9, 07745 Jena, Germany

* Correspondence: thomas.bocklitz@uni-jena.de

**Abstract**

Image registration refers to the process of spatially aligning two or more images by mapping them into a common coordinate system, so that corresponding anatomical or tissue structures are matched across images. In digital pathology, registration enables direct comparison and integration of information from different stains or imaging modalities, supporting applications such as biomarker analysis and tissue reconstruction. Accurate registration of images from different modalities is an essential step in digital pathology. In this study, we investigated how various color transformation techniques affect image registration between hematoxylin and eosin (H&E) stained images and non-linear multimodal images. We used a dataset of 20 tissue sample pairs, with each pair undergoing several preprocessing steps, including different color transformation (CycleGAN, Macenko, Reinhard, Vahadane), inversion, contrast adjustment, intensity normalization, and denoising. All images were registered using the VALIS registration method, which first applies rigid registration and then performs non-rigid registration in two steps on both low and high-resolution images. Registration performance was evaluated using the relative Target Registration Error (rTRE). We reported the median of median rTRE values (MMrTRE) and the average of median rTRE values (AMrTRE) for each method. In addition, we performed a custom point-based evaluation using ten manually selected key points. Registration was done separately for two scenarios, using either the original or inverted multimodal images. In both scenarios, CycleGAN color transformation achieved the lowest registration errors, while the other methods showed higher errors. These findings show that applying color transformation before registration improves alignment between images from different modalities and supports more reliable analysis in digital pathology.

## 1. Introduction

Image registration refers to the process of aligning multiple images within a common coordinate system, allowing for accurate spatial correspondence of anatomical or tissue structures across different acquisitions. Registration of tissue images from different modalities is an essential technique in modern healthcare, enabling significant advances in medical analysis, diagnostics, and treatment planning. By integrating data from multiple imaging modalities, clinicians and researchers can achieve more comprehensive views of tissue morphology and function. For example, in radiotherapy, the alignment of different imaging datasets is essential for accurate tumor and normal tissue definition, supporting more precise treatment approaches. Ongoing technological progress, especially in automation of image registration processes and the adoption of machine learning, continues to improve registration accuracy and workflow efficiency, directly impacting patient outcomes and enhancing radiotherapy planning and other clinical processes [1].

One area where these developments are especially important is digital pathology (DP). Accurate image registration plays a critical role in applications such as three-dimensional tissue reconstruction, integrated molecular imaging, and certain forms of detailed histological analysis [2, 3]. It allows scientists to see corresponding parts of tissue in different ways, improving understanding of structures at both the cellular and subcellular levels [4]. In addition to analyzing a single section, registering sequential or adjacent tissue sections that have been stained using different methods makes it possible to combine immunohistochemical markers and special stains in a single analysis. This approach provides greater diagnostic information than examining each stain separately and can be used as an effective alternative to chemical double staining, where tissue slices are stained individually and digitally registered to achieve similar diagnostic value [5].

Despite these benefits, achieving robust multimodal registration throughout microscopic images remains extremely challenging. Images acquired with different modalities often show substantial differences in appearance, contrast, and structural content, which makes alignment between modalities even more challenging. This problem is further complicated by non-rigid distortions introduced during histological sectioning and staining, such as holes, folding, and tearing. Additionally, artifacts like uneven lighting, dust, and bubbles are common in high-resolution imaging and can significantly impact registration accuracy [6, 7]. Beyond technical limitations, recent studies have highlighted the crucial role of accurate alignment when working with images from different modalities, especially regarding misalignment and modality mismatch. Misalignment caused by rotation, tissue deformation, or differences in sample handling can reduce the similarity between local features and make it harder for algorithms to extract and match these features effectively. Such misalignments could result either from physical rotation between serial slices or handling during slide preparation. As a result, many classical feature descriptors fail to provide robust, rotation-invariant, and stain-invariant matching, initiating the need for normalization techniques for stains [8].

To address these challenges, recent advancements in DP have led to the development of deep learning-based frameworks capable of accurately registering high-resolution, multimodal images, even under challenging contexts [3]. Cloud-based and real-time deformable registration systems have also been developed to accelerate algorithm comparison and improve whole-slide imaging workflows [6]. These innovations facilitate the wider adoption and standardization of registration of images with different modalities, allowing it to become a regular part of both research and clinical practice.

However, as imaging resolution increases and non-linear deformation effects become more significant, classical registration algorithms are often unable to provide accurate results [9]. In response, new computational approaches have emerged, including generating one modality from another using machine learning, representing images through features such as cell nuclei density, and applying segmentation-based registration with fuzzy class labeling [10].

Accurate registration is also essential for overcoming the partial volume effect and other scale-related challenges that arise when mapping high-resolution histological information onto lower-resolution modalities such as MRI [10, 11]. The precision of registration directly affects the reliability of spatial analyses, influencing both the minimum sample size required and the power of statistical models for biomarker validation [12, 13]. However, tissue deformation, heterogeneity in representation, and imaging artifacts remain ongoing challenges, sometimes leading to significant errors even when high-precision alignment is achieved [15].

Given these multifaceted challenges and the critical importance of robust registration, the optimization of preprocessing strategies remains an active area of research. Preprocessing steps such as intensity normalization, contrast adjustment, and, notably, color transformations are commonly required in an attempt to reduce both visual as well as statistical variability across modalities. These preprocessing techniques aim to improve feature correspondence and overall registration performance by addressing modality mismatch and differences in tissue representation.

In this study, we systematically evaluate the impact of different preprocessing strategies, with a focus on color transformation methods, on the registration performance between H&E-stained and multimodal images. We compare several established color transformation techniques, including Reinhard [16], Macenko [17], Vahadane [18], and CycleGAN [19], as well as a baseline without color transformation images. Registration performance is assessed using quantitative metrics such as the Median of Median relative Target Registration Error (MMrTRE), Average of Median rTRE (AMrTRE), and distances between matched key points. Our goal is to identify the color transformation technique that most effectively improves registration performance and to provide a robust framework for optimizing multimodal image integration in digital pathology.

## 2. Related Work

Medical image registration has undergone significant development, particularly in the context of DP, where aligning images from different modalities is essential for achieving a more comprehensive tissue analysis. Early methods addressed fundamental challenges such as tissue deformation, staining variability, and the complexity of gigapixel-scale images [20]. As the field advanced, more sophisticated and robust solutions have emerged to meet the demands of high-resolution, multi-modality tissue imaging.

Venet et al. [21] proposed a two-step diffeomorphic registration method that first applies affine alignment, subsequently computing an estimated non-linear deformation field to achieve precise alignment of sequentially stained histopathology slides. They also demonstrated the value of color deconvolution, which removes stain-specific color channels, minimizing artificial color differences and making registration significantly more robust. This innovation helped address one of the central challenges in aligning differently stained tissue sections.

Additionally, to address the need for automated and scalable solutions, Gatenbee et al. [22] developed the VALIS pipeline, a fully automated and open-source solution for whole-slide image (WSI) registration. VALIS employs a feature-based similarity matrix and incorporates advanced color normalization. Benchmarking results indicate that VALIS achieves state-of-the-art accuracy in WSI registration and 3D tissue reconstruction, and its flexible, user-friendly design enables integration with a wide range of imaging modalities. The adoption of color standardization in the VALIS workflow further enhances the robustness of feature matching across varying stains and imaging conditions.

Recent work by Liu et al. [20] integrated texture and spatial proximity measures into landmark registration. Their approach showed that applying color normalization prior to registration makes grayscale images more comparable, thus improving landmark detection and enabling more accurate alignment. Their methodology outperformed other registration techniques, demonstrating the importance of robust preprocessing for high-precision tasks.

Overcoming the challenges of large-angle rotation and extreme staining variation, Li et al. [8] proposed a novel framework that integrates stain-variability normalization with an orientation-free ring (OFR) feature descriptor. Their findings highlighted that transforming images into a common stain space significantly enhances feature extraction and matching, resulting in higher registration accuracy and robustness.

Despite all these advances, several important limitations persist in multimodal registration. Boehm et al. [23] and Li et al. [24] identified shortcomings related to data quality and inherent bias, while Andrews et al. [25] and Zhong et al. [26] discussed ongoing difficulties in fusing images with different spatial resolutions and the lack of standardized metrics in evaluating the quality of fusion. Kumar et al. [27] and Balluff et al. [28] noted ongoing challenges with standardized protocols, effective data integration, and incorporating new modalities, such as mass spectrometry imaging (MSI), into clinical workflows. In addition to these efforts, recent work has shown that combining complementary optical imaging modalities such as CARS, SHG, and TPEF can provide detailed biochemical and structural information. This integration leads to improved tissue discrimination and supports advanced registration approaches [29].

Across these studies, preprocessing consistently stands out as a key factor for improving registration outcomes. Schoop et al. [3] demonstrated that steps such as intensity normalization, color transformation, and denoising lead to measurable reductions in Target Registration Error and increases in Dice Similarity Coefficient. Likewise, Gatenbee et al. [22] highlighted the impact of color normalization in making feature matching more robust to staining differences, a finding also reported by Liu et al. [20] and Li et al. [8].

In summary, the field has evolved from basic affine and deformable registration algorithms to sophisticated pipelines that incorporate deep learning, stain normalization, and feature-based evaluation. Nevertheless, ensuring robust evaluation and standardization remains an open challenge, especially when integrating new preprocessing and registration strategies. This underscores the ongoing need for innovation in preprocessing and robust registration approaches to further advance the registration of images from different modalities. Our study addresses this gap by systematically comparing color transformation methods and their effect on multimodal registration performance in DP.

## 3. Materials and Methods

*3.1. Data Collection*

This study was conducted using images of 20 tissue sections taken from IBD (inflammatory bowel disease) patients, as described by Chernavskaia et al. [30] The dataset contained Crohn's disease, ulcerative colitis, and infectious colitis, with 13 male and 7 female patients aged between 20 to 65 years. Each tissue section was captured by both traditional H&E staining as well as a series of non-linear multimodal microscopy techniques. The multimodal dataset contained Coherent Anti-Stokes Raman Scattering (CARS) microscopy at two wavelengths (2850 cm$^{-1}$ and 2930 cm$^{-1}$), Two-Photon Excited Fluorescence (TPEF) microscopy across two wavelength ranges (426–490 nm and 503–548 nm), and Second Harmonic Generation (SHG) microscopy at 415 nm (LSM 510 Meta, Zeiss, Jena, Germany), offering extensive tissue information on the molecular as well as tissue structure level. After multimodal imaging, the corresponding sections were stained with H&E for scanning by brightfield microscopy, acting as the traditional histopathology reference standard. An overview of the data generation process is shown in Figure 1, illustrating both the multimodal and H&E imaging workflows.

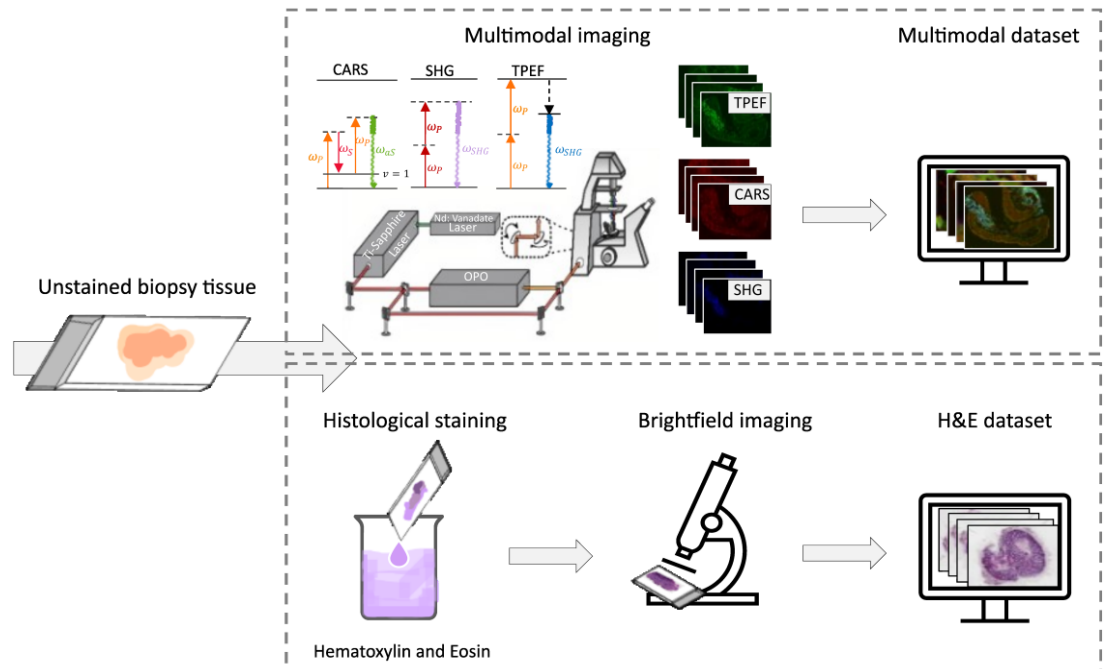

**Figure 1.** Schematic illustration of dataset generation. For each biopsy, the tissue is imaged using CARS, SHG, and TPEF microscopy to create a multimodal dataset (top row), while the same section is also stained with H&E and imaged with brightfield microscopy (bottom row) [29, 30].

*3.2. Preprocessing Techniques*

Prior to performing image registration, several preprocessing strategies were implemented to enhance feature visibility and improve the comparability between different datasets. First, contrast stretching was performed on all images to standardize their intensity distributions, highlighting tissue features and facilitating subsequent analysis. We also added a color inversion step, because in non-linear multimodal images, the highest intensity values are found in the tissue and the lowest in the background, whereas in H&E images, the relationship is reversed. This study consists of two parallel pipelines: (1)registration of H&E images with the original multimodal images, and (2) registration of H&E images with inverted multimodal images. Both pipelines used the same color transformation methods in the following steps.

We systematically explored four color transformation methods to harmonize color appearance and minimize modality differences, ensuring consistency across modalities:

- Reinhard: matches color statistics between images using channel-wise normalization in a decorrelated color space, resulting in globally consistent color adaptation [16].
- Macenko: automates stain normalization through robust separation of color channels, enabling reproducible chromatic alignment [17].
- Vahadane: decomposes image stains via structure-aware non-negative matrix factorization, transferring only color information while preserving biological detail [18].
- CycleGAN: uses adversarial learning with cycle consistency to translate images between domains, synthesizing realistic cross-modality appearances without requiring paired data [19]. For a detailed description of the CycleGAN approach, see subsection 3.2.1.

To evaluate whether these color transformation techniques improve or degrade registration, we included a baseline using images without any color transformation. The impact of these color transformation techniques is illustrated in Figure 2, which shows the results of applying color transformations on original and color-inverted multimodal images, transforming them to closely match the color characteristics of H&E images.

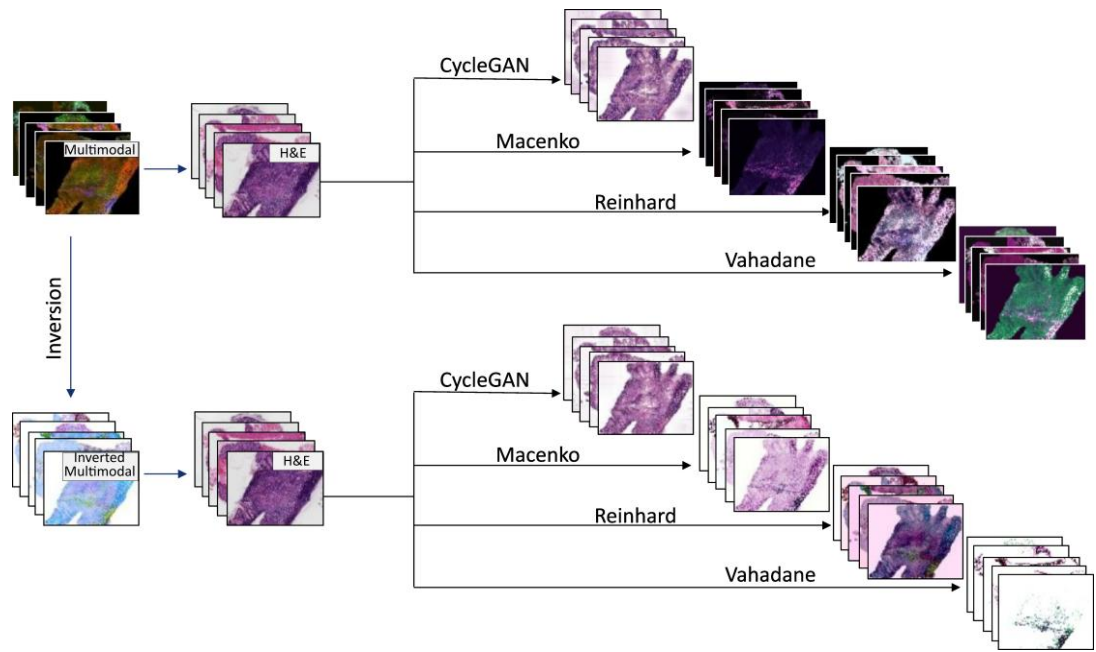

**Figure 2.** Visualization of color transformation outputs for multimodal and H&E images. The top section shows original multimodal and H&E images, followed by their output color transformed images using CycleGAN, Macenko, Reinhard, and Vahadane methods. In this case, the color transformation is applied to the multimodal images. The lower section shows the H&E and inverted multimodal images, and their color transformation outputs using the same four methods. In this case, the color transformation is applied to the inverted multimodal images.

In addition to these preprocessing strategies, the VALIS pipeline's built-in preprocessing steps were employed to further enhance image quality and alignment accuracy. Otsu masking was used to automatically segment tissue from the background, ensuring reproducible and unbiased tissue extraction. Finally, denoising was performed to enhance the clarity of key structures, facilitating improved alignment and subsequent analysis.

### 3.2.1. CycleGAN Training Details

We trained a CycleGAN for unpaired translation between H&E images and multimodal or inverted multimodal images. Each domain contained 20 images organized in separate folders. To evaluate generalization during development, we created an eighty-twenty split per domain using a random shuffle, and we used the held-out portion for validation and model selection. This procedure ensured that validation images were never seen during parameter updates.

All images were decoded, resized to 700 × 1200 pixels, and then randomly cropped to 512 × 512 patches for input to the networks. Intensities were scaled to the interval [-1, 1] using the standard linear mapping. During training, we applied spatial augmentation. For each batch, images could be rotated by 90°, 180°, or 270°, and flipped horizontally or vertically with set probabilities. Validation images underwent the same resize and crop pipeline but without augmentation, providing a consistent yet unbiased estimate of performance.

The generators follow an encoder-decoder design with skip connections and three residual blocks (instantiated with three residual connections). Instance normalization is used throughout the convolutional and transpose-convolutional layers, and the generator output uses a tanh activation to map back to the normalized intensity range. The discriminators adopt a PatchGAN architecture that produces a patchwise real-fake score map, which preserves high frequency detail (such as sharp edges and textures).

Training proceeded for 200 epochs with a batch size of one, a shuffle buffer of 100, and 200 steps per epoch. We optimized the two generators and two discriminators with Adam using a learning rate of 0.00025 and $\beta_1$ of 0.5, and random seeds were fixed for reproducibility. The overall objective combined the standard CycleGAN terms, namely adversarial loss with binary cross-entropy on logits, cycle-consistency loss with L1 distance ($\lambda = 10$), and identity loss with L1 distance weighted at half the cycle-consistency weight ($0.5\lambda$)

At the end of each epoch, generator and adversarial losses were computed on the validation set. The best model was selected based on the lowest validation loss, using early stopping with a patience of forty epochs. The weights from this best model were then used for inference.

For whole-image inference, we applied the trained generator with a sliding window strategy using 512 × 512 patches and 256 pixel overlap. Overlapping regions were blended with a Gaussian weight map to reduce visible boundaries at patch junctions. Finally, we applied bilateral filtering followed by a mild Gaussian blur, which further suppresses tiling artifacts while preserving structural detail in the reconstructed outputs.

### 3.3. Image Registration

All registrations were performed using the VALIS pipeline, which follows a sequential workflow including initial preprocessing, tissue mask generation, feature detection, and stepwise alignment using both rigid and non-rigid transformations. Rigid registration corrects global misalignments, while non-rigid registration enables local tissue deformation, together ensuring accurate alignment suitable for quantitative pathology. The original workflow is summarized in Figure 3.

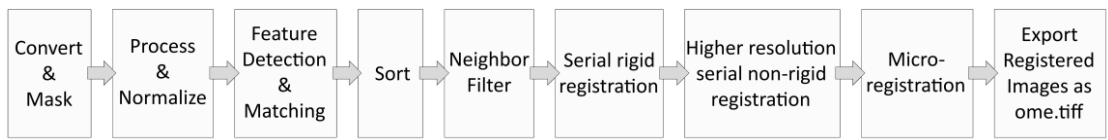

**Figure 3.** Summarized workflow of the VALIS image registration pipeline. The pipeline consists of nine main steps: image conversion and tissue masking, processing and normalization, feature detection and matching, sorting, neighbor filtering, rigid registration to correct global misalignments, higher-resolution serial non-rigid registration for local tissue deformation, micro-registration, and export of the registered images as OME-TIFF files [22].

For this study, only minimal modifications were made to the standard VALIS approach. Additional preprocessing steps were introduced, including contrast stretching, inversion, and color transformation, while sorting steps were excluded because only two images were used per sample. In addition, checkerboard visualization and custom evaluation metrics were added to support both visual and quantitative assessment of alignment performance.

### 3.4. Evaluation Metrics

Evaluation of registration performance relied on both automated and manual metrics. In the VALIS framework, landmarks are defined as biologically meaningful features, such as crypt centers or boundaries, that can be identified in both moving and reference images. These landmarks may be manually selected points, such as those provided in public benchmarking datasets like ANHIR or ACROBAT, which offer biologically significant ground truth annotations. When such manual annotations are unavailable, landmarks are derived from automatically detected and matched key points identified by the registration algorithm itself.

In VALIS, key points are typically determined using feature detection and descriptor matching methods, such as BRISK or VGG, which extract distinctive features from the images. These initial matches are then filtered for reliability before being used as landmarks for error measurement. Robust key point filtering is critical to ensure the reliability of alignment and prevent errors introduced by false matches. To improve reliability, a series of filtering strategies were applied to refine the detected key points. RANSAC (Random Sample Consensus) was used to iteratively reject mismatches by enforcing geometric consistency. This was followed by robust statistical weighting methods, such as Tukey's approach, which minimized the influence of remaining mismatches. Finally, neighborhood-based filtering ensured that the retained matches were spatially valid by considering the local arrangement of key points. These steps collectively reduced the number of key points but significantly improved the performance of the final registration. This refined set of landmarks was then used to calculate quantitative registration performance metrics.

The primary quantitative metric was the relative Target Registration Error (rTRE), defined as the Euclidean distance between corresponding landmarks normalized by the diagonal of the reference image.

$$rTRE_l^{ij} = \frac{\left\| \hat{x}_l^j - x_l^j \right\|_2}{d_j}$$

where $\hat{x}_l^j$ are the submitted coordinates of the l-th landmark in the coordinate system of image j, $x_l^j$ are the manually determined (ground truth) coordinates of the same landmark, and $d_j$ is the length of the diagonal of the reference image j [31].

To summarize registration performance across the dataset, we computed both the Median of median rTRE (MMrTRE) and the Average of median rTRE (AMrTRE). The MMrTRE represents the median of rTRE values calculated for each image pair, providing an outlier-resistant indicator of typical registration performance. The AMrTRE represents the mean of these median rTRE values, reflecting overall performance across both straightforward and challenging cases. These metrics are widely used for benchmarking registration algorithms in digital pathology studies.

In addition to the automated landmark-based evaluation, we implemented a custom point-based evaluation procedure to provide a more comprehensive assessment. For each registered image, we generated a set of ten corresponding evaluation points, manually selected across each pair of images. These points were transformed using the computed registration transformation matrices, and the Euclidean distances between each transformed point and its reference location were measured. The median of these distances was used as an additional metric of registration evaluation. The workflow for this process is shown in Figure 4.

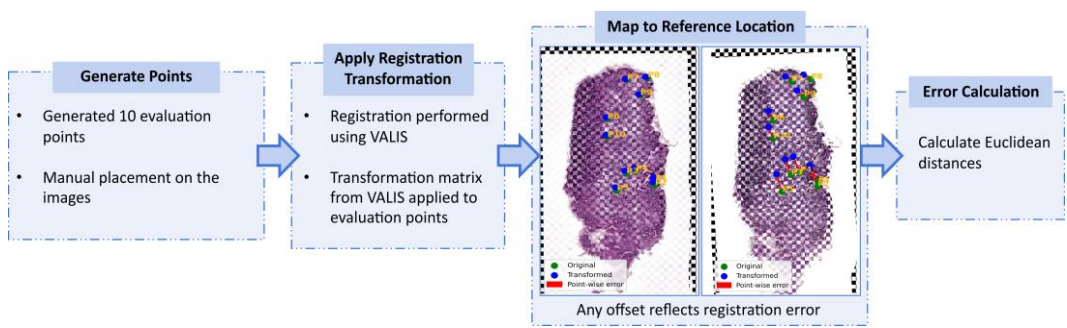

**Figure 4.** Overview of the custom evaluation metrics workflow. The first step is generating ten evaluation points per image, which are manually placed for accurate evaluation. In the second step, the registration transformation matrices from VALIS is applied to these evaluation points. The third step maps the transformed points to their reference locations, where any offset is shown as red lines indicating registration error. The final step calculates the Euclidean distances between the transformed and reference points, quantitatively evaluating the registration.

The number of matched key points identified by the VALIS method varied widely depending on the sample and transformation method. For HE and multimodal image pairs, the number of matched key points across all transformation methods and tissue samples ranged from 4 to 549 (with a median of 7 and a mean of approximately 47). For HE and inverted multimodal pairs, the values ranged from 7 to 749 (with a median of 23 and a mean of approximately 106). More key points were detected when the color similarity between the transformed image and the H&E image was higher. When the similarity was low, fewer key points were found. In contrast, the manual evaluation in this study always used exactly 10 manually selected points per sample, providing a consistent and reproducible reference for comparison.

*3.5. Implementation Details*

All analyses and image registration procedures were performed using the open-source VALIS pipeline (https://github.com/MathOnco/valis, accessed on 29 July 2025), installed according to the official documentation. The VALIS environment was set up inside a Docker container. The computational environment included Python 3.12, along with core dependencies such as pyvips, numpy, and the primary VALIS modules, as well as standard Python libraries for file and process management.

Preprocessing, registration, and evaluation steps were executed within the Docker container. Manual selection of anatomical landmarks for custom evaluation was performed using GIMP. All experiments were carried out on a workstation equipped with an AMD Ryzen Threadripper 3960X 24-core processor (48 threads, 3.79 GHz), 128 GB RAM, and dual NVIDIA GeForce RTX 3090 GPUs (24 GB GDDR6 each), running Ubuntu 22.04 with CUDA 12.6. For image-to-image translation with CycleGAN, we adapted a publicly available implementation (https://www.kaggle.com/code/virajkadam/day-night-image-translations-using-cyclegan), following the original CycleGAN paper [19].

## 4. Results

The complete workflow for image registration and evaluation is shown in Figure 5. Starting from the original hematoxylin and eosin (H&E) and non-linear multimodal tissue images, the process includes several preprocessing steps, image registration, and quantitative evaluation.

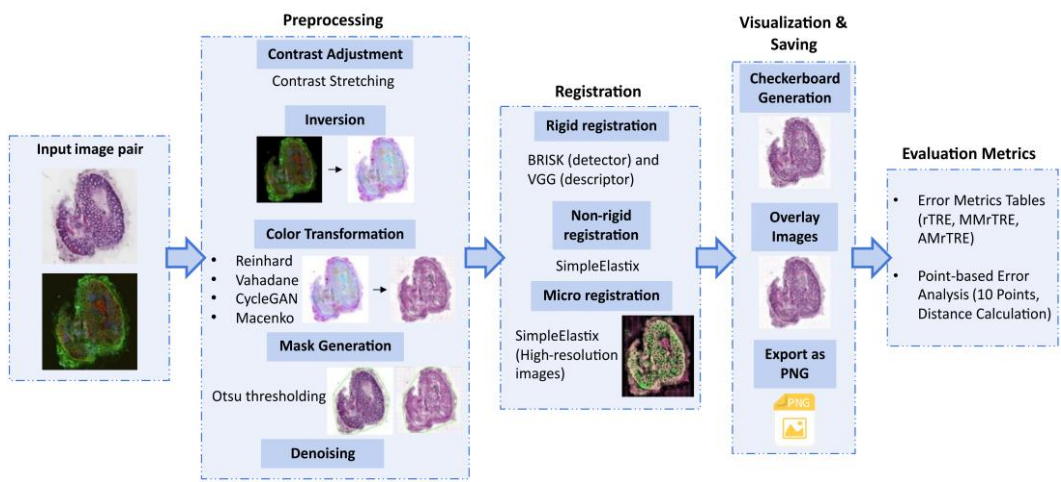

**Figure 5.** Workflow for image registration and evaluation. The first step shows the original H&E and multimodal image pairs. The second step applies a series of preprocessing techniques, including contrast adjustment, inversion, color transformation (CycleGAN, Macenko, Reinhard, Vahadane), Otsu thresholding for mask generation, intensity normalization, and denoising. The third step is image registration using the VALIS pipeline, which combines rigid alignment (BRISK detector and VGG descriptor) with a two step non-rigid registration via SimpleElastix on both low and high resolution images. The fourth step involves visualization and saving results. The final step is the quantitative evaluation, including error metrics tables and point-based error analysis using ten points per image pair.

Registration performance across the dataset was highly influenced by the choice of color transformation method. Among these methods, CycleGAN consistently achieved the best results, producing the lowest registration errors in both with and without inversion scenarios. For registrations without inversion, CycleGAN reached a median of median rTRE (MMrTRE) of 0.0088, an average of median rTRE (AMrTRE) of 0.0170, and a median point-based distance of 0.193. All other methods, including Reinhard, Macenko, Vahadane, and the baseline without color transformation, resulted in higher errors and greater variability.

Applying inversion prior to color transformation further improved registration for several methods. CycleGAN with inversion again delivered the strongest overall results (MMrTRE 0.0088, AMrTRE 0.0126, median distance 0.183). Reinhard also benefited, achieving MMrTRE 0.0086 and a median distance of 0.209, though its AMrTRE remained higher than CycleGAN (0.0486 vs 0.0126). Macenko and Vahadane showed moderate improvement with inversion but still performed less effectively overall. The baseline without color transformation improved with inversion, but did not reach the performance of CycleGAN or Reinhard.

Figure 6 shows boxplots of MMrTRE, AMrTRE, and the median point-based distance for each method, in both with and without inversion scenarios. The lowest and most consistent errors were observed for CycleGAN, particularly when combined with inversion. In contrast, Macenko, Vahadane, and the baseline without color transformation showed wider error distributions and higher median or mean values.

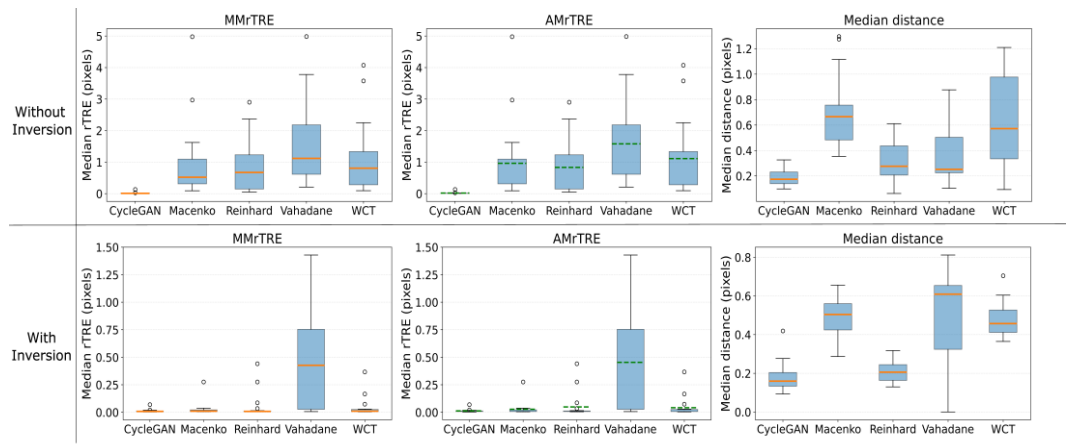

**Figure 6.** Boxplots showing evaluation metrics for image registration between different imaging modalities. Six plots are presented. The top row shows results for registration without inversion of multimodal images, while the bottom row shows results for the multimodal images with inversion. For each scenario, the left plot presents the median of median relative Target Registration Error (MMrTRE), the middle plot shows the average of median rTRE (AMrTRE), and the right plot shows the median distances from the custom point-based evaluation. WCT (without color transformation) represents the baseline method in each plot. As shown in the plots, CycleGAN provides the most consistent and accurate performance across both scenarios, while Reinhard also performs well in the with inversion scenario.

Visual inspection of checkerboard and overlay images supported the quantitative findings. CycleGAN, especially with inversion, provided close tissue boundary alignment and stable correspondence of fine structures. Reinhard with inversion produced largely coherent alignments, while Macenko and Vahadane frequently displayed local mismatches in regions with complex tissue structures. Figure 7 shows visual examples of registration results for each method, including zoomed-in areas for better comparison.

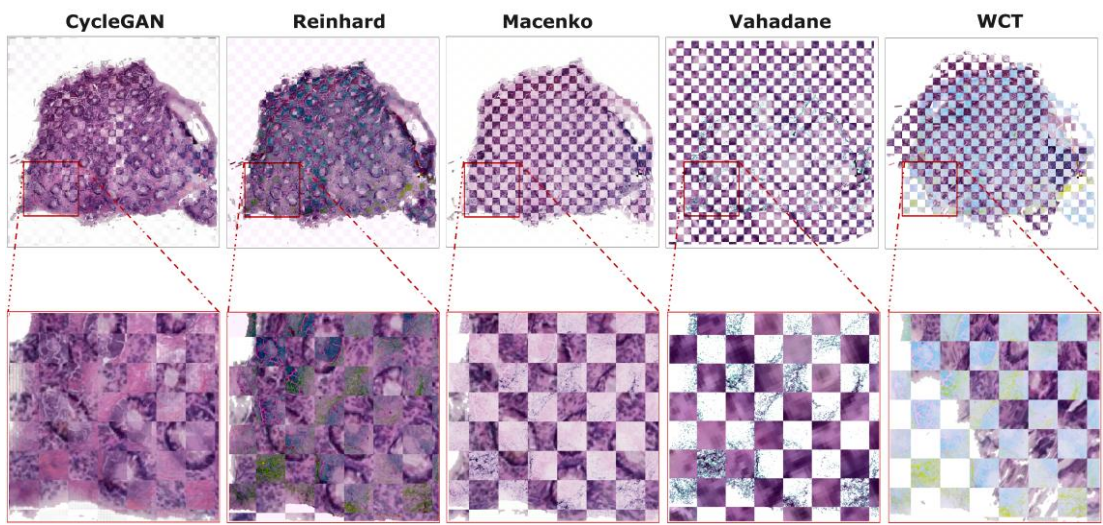

**Figure 7.** Registration results for each method on one example tissue from our dataset, with WCT indicating registration without color transformation. CycleGAN provides the closest alignment of tissue boundaries, with Reinhard performing second best. Zoomed-in regions are included to highlight differences between the methods.

## 5. Discussion

This study shows that preprocessing choices, particularly color transformation and image inversion, significantly influence the registration performance of images with different modalities. The results across all quantitative metrics highlight that applying CycleGAN color transformation, particularly in combination with inversion, leads to the most accurate and robust alignments. CycleGAN consistently produced the lowest registration errors and the least variability among tissue samples. These findings emphasize the importance of minimizing visual differences between modalities before registration, especially in digital pathology where imaging modalities often differ in appearance and background intensity.

This advantage of CycleGAN was observed consistently, as it outperformed other color transformation methods, regardless of whether inversion was applied. This performance is likely due to its ability to learn complex relationships between source and target image domains through deep learning. Traditional methods such as Reinhard, Macenko, and Vahadane rely on statistical normalization or matrix decomposition, which were less effective at reducing discrepancies between modalities in this study. For example, CycleGAN produced median registration errors well below 0.01 in both with and without inversion scenarios. Other methods, including the baseline without color transformation, frequently resulted in higher and more variable error values. The boxplot distributions were also smaller for CycleGAN, indicating more consistent performance across tissue sections.

The additional improvement observed with several color transformation methods after inversion suggests that reducing visual differences between modalities is a key to successful registration. Inverting the multimodal images made their appearance more similar to the H&E images, which facilitated more reliable detection and matching of key points during registration and resulted in better alignment both globally and locally.

These findings have practical application to digital pathology. Combination of color transformation and inversion into preprocessing pipeline can improve the registration and spatial alignment across imaging modalities. As an additional advantage, this improves downstream applications such as co-localization of molecular features, integration of serially stained slides, and fusion of structural and functional imaging. With improved registration performance, reliability of spatial and statistical analysis also increases, which is especially useful in applications such as biomarker discovery and spatial transcriptomics.

Beyond accuracy, these methods differ in what they need and where they can be used. CycleGAN learns context-dependent transformations between modalities and relies on local pixel neighborhoods during training and inference (implemented via a tiled sliding window with overlap and blending). It requires training data and greater computational resources than classical color transformation methods. To stay within resources limits, we used lower-resolution inputs for CycleGAN during both training and inference. Under these conditions, CycleGAN's per-WSI inference time was intermediate among the methods evaluated in this study.

Despite these constraints, CycleGAN offers significant advantages for multimodal image-to-image color transformation that enhance registration performance. It learns cycle-consistent mappings from unpaired data, which preserves tissue morphology and minimizes distortions during translation. CycleGAN can model nonlinear and spatially varying differences in color, contrast, and illumination, variations that classical transformation methods often cannot capture. As a result, it brings modalities to a more comparable intensity scale. This reduction in appearance differences between modalities improves feature detection and matching, and results in similarity measures that are more reliable for registration. These effects improve the robustness of the registration process. Because the transformation is learned directly from data rather than specified by predefined algorithmic rules, CycleGAN adapts to differences in staining and acquisition across slides and generalizes across samples. The use of tiled inference with overlap maintains local image context while controlling memory demands, resulting in consistent transformations for whole-slide images. Collectively, these properties reduce registration error and yield more accurate and robust alignments between modalities.

In contrast, classical color normalization methods (Reinhard, Macenko, Vahadane) are training-free. Once slide-level parameters such as LAB statistics or stain matrices are estimated, the transformation can be applied point-wise to individual pixels or even point measurements, making them more flexible for downstream analyses. They also require far fewer computational resources than deep learning methods. Because they operate in a purely point-wise manner and do not involve model training or patch-based inference, their computational cost increases linearly with image size.

Deep learning methods, in contrast, often require downsampled or patch-based inputs because the complexity of convolutional models grows nonlinearly with input resolution. As resolution grows, the number of operations and the size of intermediate feature maps expand rapidly, leading to steep increases in memory and compute demands. To handle whole-slide or gigapixel images, tiled inference with overlap is typically employed, but this strategy further increases resource consumption in order to preserve spatial context and ensure consistent transformations across tile boundaries.

In summary, CycleGAN delivers superior cross-modal harmonization at the cost of greater resource requirements and image-only applicability, while the classical methods are lighter, training-free, and point-level adaptable, but less powerful in bridging complex modality differences. The optimal choice thus depends on available data, computational resources, and the intended downstream applications. A summary of the main advantages and disadvantages of each method is presented in Table 1.

**Table 1.** Comparison of CycleGAN and classical color normalization methods in terms of data requirements, computational performance, and suitability for cross-modal registration. Key strengths and limitations are summarized for each approach.

| Method | Training Data Requirement | Runtime Performance | Field of View / Resolution Needs | Point-wise Applicability | Notes |
|---|---|---|---|---|---|
| **CycleGAN** | Yes (unpaired images, 20 per domain) | Moderate | Needs patch-based inputs because GPU memory limits prevent whole-slide inference; performance improves with richer local context (larger patches, overlap). | No | Learns nonlinear, cycle-consistent mappings; adapts to staining/acquisition variability; best harmonization across modalities |
| **Reinhard** | No (only slide statistics needed) | Moderate | Needs only enough tissue area for stable statistics; otherwise, no specific FOV limit | Yes | Simple; robust; fast to implement |
| **Macenko** | No (only stain matrix estimation) | Low | Minimal tissue area required to estimate stains; point-wise after matrix | Yes | Fast; effective for H&E normalization; widely adopted |
| **Vahadane** | No (stain matrix via NMF) | High | Needs moderate area for stable NMF; point-wise after matrix estimation | Yes | Color-faithful; slow (due to NMF); preserves stain separation |

Despite the promising results, several limitations should be considered. The dataset used in this study focused on a specific pathology, which may limit the generalizability of the findings. Future work should explore efficacy of the same preprocessing strategies in larger and more diverse datasets and in real-time clinical applications.

## 6. Conclusions

This study systematically evaluated the impact of preprocessing strategies, with a focus on color transformation methods and inversion, on the registration performance between H&E stained and multimodal images. The results demonstrated that preprocessing substantially influences registration performance, with CycleGAN consistently outperforming Reinhard, Macenko, Vahadane, and without color transformation. The combination of CycleGAN and inversion generated the most accurate and reliable alignments across all quantitative metrics, highlighting the importance of reducing visual differences between modalities before registration.

While CycleGAN delivered the strongest registration performance, it also required more training data and computational resources, as it operates on full images. In contrast, classical approaches like Reinhard, Macenko, and Vahadane are training-free and more broadly applicable, including point-based data, but were less effective at bridging substantial differences between modalities. Since image registration inherently relies on full images, the image-only limitation of CycleGAN is not a major drawback in this context, though the higher computational demand may be a practical consideration.

Future research needs to expand this evaluation to larger and more diverse datasets, such as other diseases and imaging modalities. Investigating the effectiveness of these preprocessing strategies in three-dimensional reconstruction applications and in real-time clinical environments, would make the findings more practical.

The improvements in registration performance have clear benefits for both research and clinical use. Better multimodal registration can support tasks such as combining molecular and histological features, aligning serially stained sections, merging structural and functional imaging, and analyzing biomarkers in their spatial context. By increasing the precision and reliability of multimodal integration, the proposed preprocessing strategies can help create more robust digital pathology workflows and advance spatially informed diagnostics and research.

**Author Contributions:** FD: Formal analysis, Methodology, Writing - Original Draft, Writing - Review & Editing; RE: Conceptualization, Methodology, Writing - Review & Editing, Supervision; TB: Conceptualization, Methodology, Resources, Writing - Review & Editing, Supervision, Project administration, Funding acquisition.

**Data Availability Statement:** The datasets analyzed during the current study are not publicly available but are available from the corresponding author of [30] upon reasonable request.

**Acknowledgments:** This research is supported by the German Federal Ministry of Education and Research (BMBF) through the Photonics Research Germany program and is integrated into the Leibniz Center for Photonics in Infection Research (LPI), a joint initiative of Leibniz-IPHT, Leibniz-HKI, Friedrich Schiller University Jena, and Jena University Hospital. Additional support was provided by the European Research Council (ERC). The authors gratefully acknowledge the authors of reference [30] for measuring and providing the data.

**Funding:** This work is supported by the BMBF, funding program Photonics Research Germany (13N15706 (LPI-BT2-FSU), 13N15710 (LPI-BT3-FSU), 13N15719 (LPI-BT5-FSU)) and is integrated into the Leibniz Center for Photonics in Infection Research (LPI). The LPI initiated by Leibniz-IPHT, Leibniz-HKI, Friedrich Schiller University Jena and Jena University Hospital is part of the BMBF national roadmap for research infrastructures. Co-funded by the European Union (ERC, STAIN-IT, 101088997). Views and opinions expressed are however those of the author(s) only and do not necessarily reflect those of the European Union or the European Research Council. Neither the European Union nor the granting authority can be held responsible for them.

**Conflicts of Interest:** The authors declare no conflicts of interest.